\documentclass[a4paper ]{article}

\usepackage{geometry}
\usepackage{epsfig}
\usepackage{subcaption}
\usepackage{calc}
\usepackage{amssymb}
\usepackage{amstext}
\usepackage{amsmath}
\usepackage{amsthm}
\usepackage{multicol}
\usepackage{pslatex}
\usepackage{apalike}
\usepackage[bottom]{footmisc}
\usepackage[ruled]{algorithm2e}
\usepackage{url}
\usepackage{booktabs}

\begin{document}
\title{Adaptive Combination of a Genetic Algorithm and Novelty Search for Deep Neuroevolution\thanks{Proceedings of the 14th International Joint Conference on Computational Intelligence (IJCCI 2022)}}
\author{Eyal Segal and Moshe Sipper\\[3pt]
        Department of Computer Science, Ben-Gurion University\\[3pt]
        Beer Sheva 84105, Israel\\[3pt]
        sipper@bgu.ac.il, eyalseg@post.bgu.ac.il}
\date{September 2022}
\maketitle

\begin{abstract}
Evolutionary Computation (EC) has been shown to be able to quickly train Deep Artificial Neural Networks (DNNs) to solve Reinforcement Learning (RL) problems. While a Genetic Algorithm (GA) is well-suited for exploiting reward functions that are neither deceptive nor sparse, it struggles when the reward function is either of those. To that end, Novelty Search (NS) has been shown to be able to outperform gradient-following optimizers in some cases, while under-performing in others. We propose a new algorithm: Explore-Exploit $\gamma$-Adaptive Learner ($E^2\gamma AL$, or EyAL). By preserving a dynamically-sized niche of novelty-seeking agents, the algorithm manages to maintain population diversity, exploiting the reward signal when possible and exploring otherwise. The algorithm combines both the exploitation power of a GA and the exploration power of NS, while maintaining their simplicity and elegance. Our experiments show that EyAL outperforms NS in most scenarios, while being on par with a GA---and in some scenarios it can outperform both. EyAL also allows the substitution of the exploiting component (GA) and the exploring component (NS) with other algorithms, e.g., Evolution Strategy and Surprise Search, thus opening the door for future research.
\end{abstract}

\section{\uppercase{Introduction}}

As the field of Reinforcement Learning (RL) \cite{sutton2018reinforcement} is being applied to harder tasks, two unfortunate trends emerge: larger policies that require more computing time to train, and ``deceptive'' optima. While gradient-based methods do not scale well to large clusters, evolutionary computation (EC) techniques have been shown to greatly reduce training time by using modern distributed infrastructure \cite{arxiv.1703.03864,arxiv.1712.06567}.

The problem of deceptive optima has long since been known in the EC community: Exploiting the objective function too early might lead to a sub-optimal solution, and attempting to escape it incurs an initial loss in the objective function. Novelty Search (NS) mitigates this issue by ignoring the objective function while searching for new behaviors \cite{Lehman2008}. This method had been shown to work for RL \cite{arxiv.1712.06567}. 

While both genetic algorithms (GAs) and NS have been shown to work in different environments \cite{arxiv.1712.06567}, we attempt herein to combine the two to produce a new algorithm that does not fall behind either, and in some scenarios surpasses both. 

Section~\ref{sec:prev} presents previous work. 
In Section~\ref{sec:methods} we describe the methods used herein: GA, NS, and our new algorithm. 
The experimental setup is delineated in Section~\ref{sec:exp}, 
followed by results in Section~\ref{sec:results}. 
We end with a discussion in Section~\ref{sec:disc} and future work in Section~\ref{sec:future}.

\section{\uppercase{Previous Work}}
\label{sec:prev}
\subsection{Reinforcement Learning}
In Reinforcement Learning (RL) problems the goal is to find a behavior that accomplishes a specific task, without explicitly telling the learner what to do; instead, a given reward function judges the behavior of the learner. The learning algorithm's purpose is to devise a behavior that will maximize the cumulative returns of the given reward function \cite{sutton2018reinforcement}.

\subsection{Evolutionary Computation in Reinforcement Learning}
While most traditional RL research attempts to assign credit to previous actions and diffuse the rewards throughout the run \cite{sutton2018reinforcement}, EC attempts to holistically judge the agent as a black box: The cumulative sum of episodic rewards is used as a fitness metric, without assigning credit to intra-episode actions. While these methods can sometimes be less sample-efficient, they scale well with the number of CPUs available, and thus save wall-clock time overall \cite{arxiv.1703.03864,arxiv.1712.06567}.

Evolution Strategies (ES) \cite{rechenberg1989evolution,beyer2002evolution} work by empirically estimating the gradient. By sampling enough points near the current solution, one can estimate the impact of ``nudging'' the solution to any direction. When applied to RL, ES can be used to adjust the weights of a DNN-based policy to maximize rewards, as shown in \cite{arxiv.1703.03864}.

The Genetic Algorithm (GA) is a gradient-free algorithm that works by allowing higher-fitness agents to reproduce and accrue random mutations. Novelty Search (NS) works similarly, but instead of being driven by fitness, it disregards the reward function and searches for novel behaviours. GA and NS were shown to be able to solve RL problems in \cite{arxiv.1712.06567}.

\subsection{Exploration-Exploitation in Evolutionary Computation}
While a GA and Novelty Search both optimize either fitness or novelty (but not both), previous attempts have tried to optimize both simultaneously. 

\paragraph{Multi-Objective Optimization}
Several algorithms have previously managed to simultaneously optimize more than one objective, by looking for solutions on the Pareto front. The basic idea is to hold a number of such solutions, where each one represents some trade-off between the objectives \cite{1597059}. These methods do not apply to the domain of RL: though the exploitation (sum of rewards) can be a dimension in the Pareto front, the exploration (novelty) cannot. The latter is because the goal of exploration is dynamic---once a solution is found to be dominating, it is no longer novel and thus by definition not on the Pareto front. Additionally, the objective is not to find a novel solution that is also well-adjusted (and vice versa)---novelty is just a mechanism to explore the search space more effectively.

\paragraph{SAFE} 
In the SAFE algorithm (Solution And Fitness Evolution) a population of problem solutions coevolves along with a population of objective functions, wherein each objective function unifies all objectives into a single objective \cite{sipper2019solution,sipper2019multiobjective}. The objective functions' evolution is driven by genotypic novelty, which does not suit our purpose---we would like to increase or decrease the importance of novelty as the population becomes less or more diverse, respectively---and not try new ratios independently. 

\paragraph{Method II}
A previous attempt to hybridize a GA and NS was done in \cite{jackson2019novelty}. In their so-called Method II, whenever the GA's population grew stale---i.e., no improvement in cumulative rewards---the population was resampled from the NS's archive instead, looking for agents with behavior as different from the current, stale population as possible. In some simulations, Method II managed to avoid local optima, which the GA did not manage to escape, thus continuing to learn and achieve better results. 

\paragraph{Quality-Diversity}
In Quality-Diversity methods such as Novelty Search with Local Competition \cite{lehman2011evolving} and MAP-elites \cite{mouret2015illuminating,colas2020scaling}, diversity is ensured by dividing the population into cells (niches), and quality is ensured by allowing local competition within each cell. 

\section{\uppercase{Methods}}
\label{sec:methods}
In this section we present an outline of the neuroevolution policy representation, 
and the pseudo-code of the GA used (Algorithm~\ref{alg:ga}) and of NS (Algorithm~\ref{alg:novelty}).
Then, we present our technique for hybridizing them in Algorithm~\ref{alg:aee}.

\subsection{Policy Representation}
\label{method:representation}
In order to find a policy through neuroevolution, a representation scheme for the policy is needed. One such representation is a pre-configured, fixed-architecture DNN whose weights are encoded as a fixed-length vector. The architecture can also be encoded and evolved using NEAT \cite{stanley2002evolving} or one of its later variants. Decision Trees can be encoded through Genetic Programming \cite{koza1992evolution}.

\subsection{Genetic Algorithm}
The GA (Algorithm~\ref{alg:ga}) \cite{holland1992genetic} is a population-based algorithm, in which the population is measured against a fitness function. At each iteration, the better-fitted individuals survive while the lesser-fitted ones die off. The survivors then reproduce, and their descendants mutate to accrue random mutations---which explore the search space. In the domain of RL the fitness function is usually the sum of returns of the reward function.

\begin{algorithm}
\SetAlgoLined
\caption{Genetic Algorithm}
\label{alg:ga}
\SetKwInput{Input}{input}
\Input{max training steps, popsize}

\While{training step $<$ max training steps}{
\eIf{first generation}
{population $\leftarrow$ initialize(popsize)}
{
survivors $\leftarrow$ select survivors(population, fitnesses) \\
parents $\leftarrow$ select parents(survivors, popsize - 1) \\
children $\leftarrow$ mutate(parents) \\
population $\leftarrow$ children + \{generation elite\} \\
}\vspace{\baselineskip}

trajectories $\leftarrow$ rollouts(population) \\
fitnesses $\leftarrow$ sum rewards(trajectories)\\
\vspace{\baselineskip}
generation elite $\leftarrow$ extract elite(population, fitnesses) \\
report elite(generation elite)
}

\end{algorithm}

\subsection{Novelty Search}

\begin{algorithm}
\SetAlgoLined
\caption{Novelty Search}
\label{alg:novelty}
\SetKwInput{Input}{input}
\Input{max training steps, popsize}

\While{training step $<$ max training steps}{
\eIf{first generation}
{population $\leftarrow$ initialize(popsize)}
{
survivors $\leftarrow$ select survivors(population, novelty scores) \\
parents $\leftarrow$ select parents(survivors, popsize - 1) \\
children $\leftarrow$ mutate(parents) \\
population $\leftarrow$ children + \{generation elite\} \\
}\vspace{\baselineskip}

trajectories $\leftarrow$ rollouts(population) \\
fitnesses $\leftarrow$ sum rewards(trajectories)\\
\vspace{\baselineskip}

generation elite $\leftarrow$ extract elite(population, fitnesses) \\ 
report elite(generation elite)
\vspace{\baselineskip}

bcs $\leftarrow$ behavior characterstic(trajectories) \\
novelty scores $\leftarrow$ novelty measure(bcs, archive) \\

\vspace{\baselineskip}
update archive(bcs)
}
\end{algorithm}

In some cases, the reward signal can be deceptive. For example, in the short run, waiting for an elevator will get you no closer to the 100th floor and will yield no rewards, while climbing the stairs will grant immediate rewards. Thus, the reward function deceives you into taking the stairs, while waiting a few minutes for the elevator will get you closer to the objective in the long run. To that end, Lehman and Stanley \cite{Lehman2008} presented \textit{Novelty Search} (Algorithm~\ref{alg:novelty}), which essentially ignores the objective and searches for behavioral novelty (using a novelty metric that requires careful consideration).

Novelty Search  is an exploratory algorithm, but when it gets ``close'' to an optimum it does not always optimize towards it before exhausting the rest of the search space.

\subsection{Adaptive Explore-Exploit}
EyAL (Algorithm~\ref{alg:aee}) combines the two algorithms---one mainly exploratory, the other mainly exploitative---and introduces a control variable $\gamma$ to control the size of the exploratory niche. We have chosen the Genetic Algorithm to be the exploitative algorithm, and Novelty Search to be the exploratory algorithm. It should be noted, however, that these can be easily swapped with other algorithms, which we leave for future research.

At every generation, both the fitness (sum of rewards) and novelty score of each specimen in the population is calculated. Then, $(\text{population size}) \times \gamma$ children are created using the novelty scores, and $(\text{population size}) \times (1 - \gamma)$ are created using the fitness. This ensures the survival of two niches: one that tries to optimize the rewards, and one that tries to search for new, unexplored behaviours. 

The main principle in this approach is that $\gamma$ is dynamic, and auto-adjusts during the run. If the population grows stale (no improvement in fitness with relation to the previous generation), $\gamma$ increases to promote exploration and increase the size of the exploring niche. Otherwise, $\gamma$ decreases to promote exploitation, increasing the size of the exploiting niche. 








\begin{algorithm}
\SetAlgoLined
\caption{Explore-Exploit $\gamma$-Adaptive Learner}
\label{alg:aee}
\SetKwInput{Input}{input}
\Input{max training steps, population size, initial exploration rate $\gamma$, exploration growth rate $\alpha$, exploration decay rate $\beta$}

\While{training step $<$ max training steps}{
\eIf{first generation}
{population $\leftarrow$ initialize(popsize)}
{
exploring survivors $\leftarrow$ select survivors(population, novelty scores) \\
exploiting survivors $\leftarrow$ select survivors(population, fitnesses) \\
\vspace{\baselineskip}
exploring parents $\leftarrow$ select parents(exploring survivors, $\gamma$ $\times$ (popsize - 1)) \\
exploiting parents $\leftarrow$ select parents(exploiting survivors, (1 - $\gamma$) $\times$ (popsize - 1) ) \\
\vspace{\baselineskip}

children $\leftarrow$ mutate(exploring parents) + mutate(exploiting parents) \\
population $\leftarrow$ children + \{generation elite\} \\
}

\vspace{\baselineskip}
trajectories $\leftarrow$ rollouts(population) \\
fitnesses $\leftarrow$ sum rewards(trajectories)\\
\vspace{\baselineskip}

generation elite $\leftarrow$ extract elite(population, fitnesses) \\ 
report elite(generation elite)
\vspace{\baselineskip}

\eIf{generation elite $>$ previous generation elite}
{$\gamma \leftarrow \gamma - \beta$}
{$\gamma \leftarrow \gamma + \alpha$}
$\gamma \leftarrow clamp(0, 1, \gamma)$ \\
\vspace{\baselineskip}

bcs $\leftarrow$ behavior characteristic(trajectories) \\
novelty scores $\leftarrow$ novelty measure(bcs, archive) \\

\vspace{\baselineskip}
update archive(bcs)
}
\end{algorithm}

\subsection{Operators and Terminology} \label{operators}
While the pseudo-code describes the general scheme, the details have been left out. This design allows modularity---one can easily swap the implementation of any of these operators with ease. We believe that this design both promotes the elegance and clarity of the algorithm itself and allows easy experimentation with different operator implementations. In this section we present the operators used in the algorithm and provide information about our implementation for said operators.

\paragraph{Policy}
A policy is a strategy that allows an agent to make decisions; at each state, the policy determines which action should be taken. We have used a fixed-architecture deep neural network (DNN) to represent our policy; additional possibilities for such encodings are discussed in \ref{method:representation}. 

In discrete environments, the (DNN) represents a state-value function, taking an observation from the environment and returning a vector of possible actions and their respective (internal) values. The agent then chooses the action with the highest value (i.e., in a deterministic manner---probabilistic selection is also possible, although not in the scope of this work).

In continuous environments the output represents a single action. For example, if the task is controlling the $k$ joints of a robot, the output is a $k$-dimensional vector.

\paragraph{Trajectory}
A trajectory represents a single, complete run of an agent in the environment, consisting of a chain of
$<$state, action, reward$>$ tuples.

\paragraph{Rollout}
A rollout takes an agent, performs a single run in the environment, and returns the trajectory of the run.

\paragraph{Initializer}
The initializer method creates initial agents for the evolutionary algorithm. In our context, this method generated the parameter vectors for the agent policy DNNs. We used the default initialization in PyTorch \cite{paszke2019pytorch}.

\paragraph{Survivor Selection}
This operation receives the population and the fitness of each individual in the population, and returns the subset of the population that survived this generation. We used truncated selection, which selects only the \textit{truncation size} specimens with the highest fitness.

Other methods that were not used in this paper include the selection of only the newest solutions, the selection of only the elite, and more. 

\paragraph{Parent Selection}
An operation that receives survivors and the number of parents to output, and selects this number of parents from the survivors. We used random selection with repetition. Since we did not perform crossover, we selected a single parent per specimen in the population.

\paragraph{Mutation}
This method receives an agent, and returns a slightly mutated agent. We added a random Gaussian noise vector $\overrightarrow{v} \thicksim \mathcal{N}(0, \sigma^2)$ to the DNN's parameter vector, where $\sigma^2$ (Mutation Power) is a hyperparameter.

\paragraph{Extract Elite}
This method receives the population and the matching fitness of each individual in the population, and returns the generation's elite. Because the environments are stochastic, we took the \textit{elite candidates} individuals with the highest fitness and tested them against \textit{elite robustness} more rollouts (these were counted towards the algorithm's training steps). The agent with the highest average score in these rollouts was chosen as the elite. Both \textit{ elite candidates} and \textit{elite robustness} are hyperparameters.

\paragraph{Behavior Characteristic}
This method receives a trajectory and returns a vector (preferably shorter than that of the trajectory) that represents an aspect of the trajectory. Novelty is measured with respect to that characteristic. The behavior characteristic is domain-specific; we used the following in our experiments: use the last observation vector of the trajectory, and concatenate the last time-step (the length of the trajectory chain) normalized by the maximal \textit{timestep} allowed in the environment.

\paragraph{Novelty Measure}
This method receives the current population---represented by their respective \textit{behavior characteristics}---and an \textit{archive} of previous generations, and measures how \textit{novel} each specimen of the population is (with respect to the \textit{behavior characteristics}). We used the average distance from the \textit{k-nearest-neighbors} as a novelty measure, where \textit{k} is a hyperparameter.

\paragraph{Update Archive}
This method updates the archive to contain some representation of the current population. Since we do not wish to store the entirety of the evolution, we only store each individual of the population with \textit{pr} probability, where \textit{pr} is a hyperparameter.

\paragraph{The code} is available at \url{github.com/EyalSeg/ecrl}.

\section{\uppercase{Experiments}}
\label{sec:exp}
\subsection{Trials}
A \textit{trial} is an independent, complete run of the algorithm in a specific environment. 

When a generation finishes, the elite reported by the algorithm is evaluated for another 100 validation episodes. This evaluation is used for reporting purposes only, and the algorithm does not get to use these results; as such, they do not count towards the time-step limit. The \textit{score} of a trial is the highest validation score of an elite (from \textit{any} generation). 

We created our own maze environment in Mujoco \cite{todorov2012mujoco}, in which we ran 40 trials of the tested algorithms.

\subsection{Hyperparameters}
As with many optimization algorithms, evolutionary algorithms require hyperparameters, and ours is no exception. We used the same hyperparameters in all algorithms (where applicable). Due to a shortage of computational resources available to us, the values for the hyperparameters were derived through limited trial-and-error experimentation (in the future we plan to perform a more thorough hyperparameter sweep, resources permitting). 
Table~\ref{tab:hyper} lists the full set of hyperparameters.

The network architecture---comprising two 256-units linear layers and $tanh$ activation---is that used by \cite{arxiv.1712.06567}.

\subsection{Mujoco Maze}
\begin{figure}
  \centering
  \includegraphics[scale=0.33]{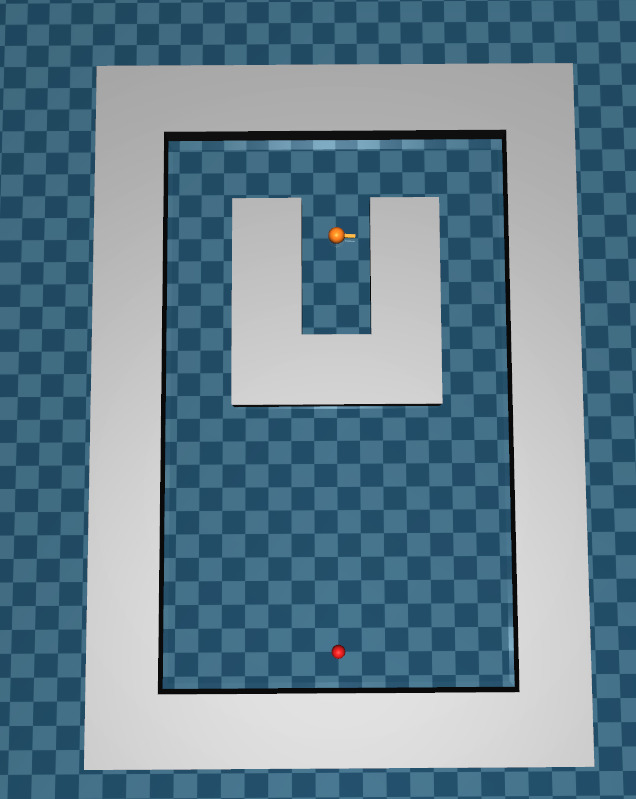}
  \caption{Maze environment. The agent is the orange dot, the destination is the red dot. Observe the obstacle creating a pocket in between the initial position of the agent and the destination.}
  \label{fig:maze}
\end{figure}

We hypothesized that $EyAL$ will perform better in environments in which the reward itself can deceive the agent vis-a-vis the objective; thus, greedily optimizing the reward will lead to poor episodic rewards overall. To test this we created a simple maze (Figure~\ref{fig:maze}) in which going straight toward the exit will lead to an obstacle. To circumvent this, the agent must go back---\textit{away} from the destination---which incurs a penalty.

We tested the algorithms in this environment with respect to two reward functions.  In \textit{PointMazeDeceptive}, at every time step the agent receives a negative reward equal to the Euclidean distance from the agent's position to the destination. When the agent gets to the exit it receives a positive reward of 10,000 points; thus, if the agent gets to the destination it will have a positive episodic reward.
In \textit{PointMazeSparse}, at every time step the agent receives a negative reward of $-1$, and thus the episodic score will be the negative of the number of time steps it took the agent to reach the exit, or $-500$ if it did not escape.

The behavior characteristic for this environment is defined as the last position the agent had been in.

\begin{table}
  \caption{General hyperparameters. Some are represented by a symbol, shown in parentheses.}
  \label{tab:hyper}
  \centering
  \begin{tabular}{rl}
    \toprule
    Designation & Value\\
    \midrule
    DNN Hidden Layer Dimensions & 256, 256 \\
    DNN Activation & tanh \\
    Mutation Power ($\sigma^2$) & 0.005 \\
    Population Size (M) & 50 + 1 \\
    Truncation Size (T) & 20 \\
    Elite Candidates & 5 \\
    Elite Fitness Robustness & 5 \\
    Novelty Archive Probability (pr) & 0.01 \\
    Novelty K-Nearest Neighbors (K) & 25 \\
    Exploration Growth Rate ($\alpha$) & See Table~\ref{tab:hyper2} \\
    Exploration Decay Rate ($\beta$) & See Table~\ref{tab:hyper2}  \\
    (Initial) Exploration Rate ($\gamma$) & See Table~\ref{tab:hyper2} \\
    Training Steps & See Table~\ref{tab:hyper2} \\
    Validation Episodes & 100 \\
  \bottomrule
\end{tabular}
\end{table}

\begin{table}
\caption{Environment-specific hyperparameters: $\alpha$, $\beta$, $\gamma$, and training steps.}
\label{tab:hyper2}
\centering
\begin{tabular}{rcccc}
\toprule
Environment & $\alpha$ & $\beta$ & $\gamma$ & Steps \\ 
\midrule
PointMazeSparse & 0.1 & 0.1 & 0.75 & $5e^7$ \\ 
PointMazeDeceptive & 0.1 & 0.1 & 0.75 & $5e^7$ \\ 
\bottomrule
\end{tabular}
\end{table}

\section{\uppercase{Results}}
\label{sec:results}
Comparing different evolutionary algorithms is often not straightforward, and care must be taken to compare in a fair manner (e.g., account for differences in resources expended). 
Since our score reporting is done at the end of each generation, the fitness-vs-timestep curve is sparse---which makes aggregating multiple runs complicated, as not all fitness reports share the same timestep. To allow such an aggregation, we split the x-axis (time steps) into intervals, and used ``forward-fill'' to populate the figure at these interval points: by propagating results forward from previous time steps, we ensure that at these interval points the algorithm accounts for the previously known best result. This method allows the comparison of all runs at the same points, enabling the plotting of a fair mean and confidence bounds. 

\begin{figure}
  \centering
  \includegraphics[scale=0.65]{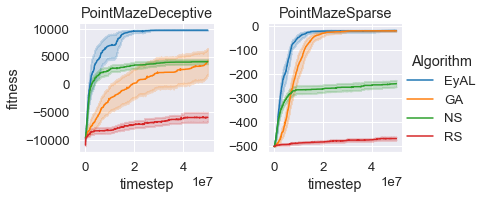}
  \caption{Comparison of EyAL, GA, NS, and Random Search (RS) in our custom Mujoco environment. Plots include the mean of 40 trials, along with 95\% confidence intervals. 
  Since every generation takes a different number of time steps, comparisons were made every 10,000 time steps, using forward-fill from previous generations.
  }
  \label{fig:results_maze}
\end{figure}

Figure~\ref{fig:results_maze} presents the results of our experiments in our custom Mujoco environment. 


\section{\uppercase{Discussion}}
\label{sec:disc}
While our computational resources were rather limited, allowing much smaller populations and far less time than \cite{arxiv.1712.06567}, EyAL managed to achieve good results. 




In PointMazeSparse, the reward signal is very sparse: for every time step the robot did not exit the valley the agent gets a negative reward of -1. Thus, until it finds a viable solution, the learner does not receive any signal it can exploit. In this regard, a specimen that is just one mutation away from finding the exit is just as unfit as an agent that does not even try to exit. This fact makes this problem hard for eager-exploiters such as a GA. On the other hand, when an agent gets to the exit once, any other solution that finds the exit is no longer novel, which makes it harder for NS to obtain a solution that gets to the exit slightly faster. When the novelty-seeking niche of EyAL finds the exit, the episodic reward increases---and the exploiting niche increases in size. Thus, while NS seeks new positions it has never been to (which are not viable solutions), EyAL optimizes the solution it found---which explains why EyAL improves a bit even after NS starts to plateau . 

In PointMazeDeceptive, the reward signal deceives the learner to run into the obstacle. For a GA, any specimen that goes back from the obstacle and then does not proceed straight down, will incur a negative penalty, and will not survive to the next generation. To escape the obstacle, a single random mutation has to make the agent both go around the obstacle and then go downwards. For NS, however, just going around the corner yields an immediate improvement. But, when the destination is reached, NS has a hard time improving it, because novelty lies in exhausting other destinations. While it does take EyAL slightly longer to find the destination (with respect to NS), it optimizes much better when the destination is found---which explains why EyAL keeps improving after NS plateaus.

\begin{figure}
  \centering
  \includegraphics[scale=0.5]{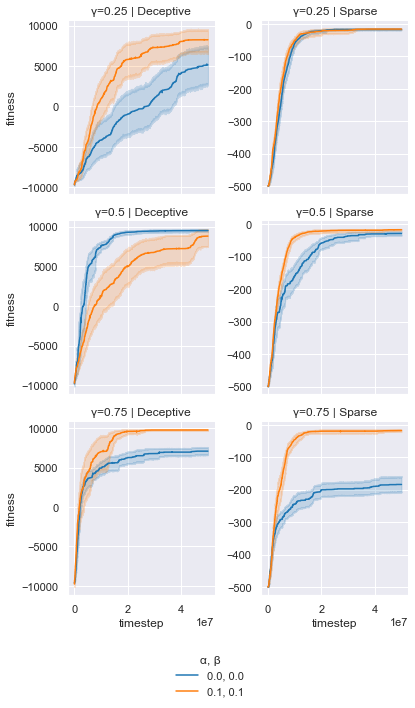}
  \caption{Comparison of an adaptive ($\alpha \neq0, \beta \neq 0$) and an un-adaptive ($\alpha = 0 = \beta$) Explore-Exploit learners. Plots include the mean of 20 trials, along with 95\% confidence intervals.
  }
  \label{fig:results_ab}
\end{figure}

In Figure~\ref{fig:results_ab}, the benefit of an adaptive $\gamma$ can be seen. In the deceptive case, the adaptive version is best at most initial $\gamma$ values tested, with the exception of one. In the sparse setting, the adaptive version consistently improves the results---or at least does not harm them. 

One interesting finding is the difference between the algorithms in the two PointMaze environments tested. Unsurprisingly, NS is not affected by the choice of reward function, as it does not try to optimize for it. On the other hand, GA suffers greatly from the deceptive function. While EyAL does suffer from deception, the effect of it is not as adverse as it is for GA. In the sparse environment, EyAL learns faster than GA but does not provide a better solution overall. In the deceptive environment, EyAL---which hybridizes GA and NS---provides a better solution than either. We find this fact interesting, as adding the under-performing GA to NS improves NS.

\section{\uppercase{Future Work}}
\label{sec:future}
\paragraph{Optimizing EyAL}
The modular design of EyAL allows for the replacement of many genetic operators, as mentioned in \ref{operators}. Likewise, the underlying exploring and exploiting algorithms can also be modified with different flavours of GA (such as introducing fitness sharing \cite{mckay2000fitness}, self-adaptive mutation \cite{schwefel1981numerical}, or replaced altogether by algorithms such as Evolution Strategies or Surprise Search. As the core idea of this paper was to explore whether the dynamic hybridization of two algorithms can result in a third, better algorithm---these ideas are left for future work. 

It should be mentioned that our implementation of behaviour characteristic was naive, yet even with this basic BC our technique yielded improvement over both the standard GA and NS. 

\paragraph{EyAL and Quality-Diversity}
Since we did not optimize EyAL, we left direct comparison to state-of-the-art QD methods for future work.

It should be mentioned that the principles of QD and of EyAL are not mutually exclusive. While QD methods use a fixed number of niches, the adaptiveness of EyAL can be introduced to increase and decrease the number of cells, or to allocate additional computational resources to more promising niches at the expense of less promising niches. Likewise, the local-competition principles of QD can be introduced to EyAL by various methods of fitness sharing \cite{mckay2000fitness}. 

While the global competition of EyAL has been shown to be inferior to local competition in \cite{colas2020scaling}, the adaptiveness of EyAL is yet to be explored in this context. An algorithm that exploits both of these traits would be interesting to see.

\small
\bibliographystyle{apalike}
\bibliography{bib}

\end{document}